\documentclass[conference]{IEEEtran}
\usepackage{cite}
\usepackage{amsmath,amssymb,amsfonts}
\usepackage{algorithmic}
\usepackage{graphicx}
\usepackage{textcomp}
\usepackage[frozencache=true,cachedir=minted-cache]{minted}
\usepackage[usenames,dvipsnames]{xcolor}
\usepackage[all]{tcolorbox}
\usepackage{tabularx}
\usepackage{array}
\usepackage{colortbl}
\usepackage{caption}
\usepackage{subcaption}

\tcbset{tab2/.style={enhanced,fonttitle=\bfseries, fontupper=\small\sffamily,
colback=blue!5!white,colframe=blue!50!black,colbacktitle=cyan!40!white,
coltitle=black,center title}}

\begin{document}

\title{RF Fingerprinting Needs Attention: Multi-task Approach for Real-World WiFi and Bluetooth
\vspace{-0.3cm}}

\author{\IEEEauthorblockN{Anu Jagannath${^{\ddagger\dagger}}$, Zackary Kane$^\ddagger$, Jithin Jagannath${^{\ddagger \aleph}}$}
\IEEEauthorblockA{ $^{\ddagger}$Marconi-Rosenblatt AI/ML Innovation Lab, ANDRO Computational Solutions, LLC, Rome NY \\ 
$^{\dagger}$Institute of Wireless Internet of Things, Northeastern University, Boston, MA \\
$^{\aleph}$ University at Buffalo, Buffalo, NY,  Email: \{ajagannath, zkane, jjagannath\}@androcs.com
\vspace{-0.5cm}
}}
\maketitle

\begin{abstract}
A novel cross-domain attentional multi-task architecture - xDom - for robust real-world wireless radio frequency (RF) fingerprinting is presented in this work.
To the best of our knowledge, this is the first time such comprehensive attention mechanism is applied to solve RF fingerprinting problem. In this paper, we resort to real-world IoT WiFi and Bluetooth (BT) emissions (instead of synthetic waveform generation) in a rich multipath and unavoidable interference environment in an indoor experimental testbed. We show the impact of the time-frame of capture by including waveforms collected over a span of months and demonstrate the same time-frame and multiple time-frame fingerprinting evaluations. The effectiveness of resorting to a multi-task architecture is also experimentally proven by conducting single-task and multi-task model analyses. Finally, we demonstrate the significant gain in performance achieved with the proposed xDom architecture by benchmarking against a well-known state-of-the-art model for fingerprinting. Specifically, we report performance improvements by up to 59.3\% and 4.91$\times$ under single-task WiFi and BT fingerprinting respectively, and up to 50.5\% increase in fingerprinting accuracy under the multi-task setting.
\end{abstract}

\begin{IEEEkeywords}
RF fingerprinting, temporal attention, spatio-temporal attention, time-frequency attention. WiFi, Blueetooth
\vspace{-0.5cm}
\end{IEEEkeywords}

\section{Introduction and Related Works}
A recent statistical survey forecasts more than 2 billion WiFi 6 device shipments by the end of 2021 (i.e., over 50\% of all WiFi shipments) and that number will grow to 5.2 billion devices by 2025 \cite{qualcomm}. According to a statistical report by Statista, annual Bluetooth (BT) device shipments worldwide stood at 4.7 billion units in 2021 \cite{stat}. The threat surfaces and privacy concerns exacerbated by such accelerated device roll-outs especially in densely populated areas such as hospitals, airports, Federal security centers, etc., 
is unprecedented.

Wireless radio frequency (RF) fingerprinting is emerging as an easy-to-deploy security scheme to counter and mitigate the security issues arising from having unknown or prohibited RF emitters sharing the wireless spectrum \cite{RFfingerChallenges,Ajagannath2022ComST2022}. \textit{RF fingerprinting} refers to the identification of wireless transmitters from their received emissions based on inadvertent features embedded in their waveforms. These inadvertent features emerge from their internal RF circuitry. Due to the imperfections entailed in the manufacturing process, the components of the RF circuit such as the power amplifier (PA), low noise amplifier (LNA), clock circuits, and local oscillators (LO), among others introduce IQ imbalance, clock skew, out of band (OOB) spurious leakage, which differs across the devices even by the same manufacturer. These minuscule but unique features form the RF signature of the wireless device which with careful signal processing can be extracted without any prior knowledge or induced impairments, by what is referred to as a \textit{truly blind RF fingerprinting}. Extracting these faint RF signatures from commercial-off-the-shelf (COTS) Internet of Things (IoT) devices from the same family in a blind manner is a daunting task. Recent advances in deep learning have made a significant impact on the wireless domain \cite{JagannathAdHoc2019, AJagannath22PHYCOM, sankhe2019oracle,shawabka2020exposing, Ajagannath6G2020}. The effectiveness of deep learning especially convolutional neural network (CNN) in extracting unique device relevant features from the RF emissions has led to an active research area in this direction \cite{Ajagannath2022ComST2022,sankhe2019oracle,shawabka2020exposing}.

A CNN-based framework inspired by AlexNet \cite{alexnet} for RF fingerprinting called ORACLE (Optimized Radio clAssification through Convolutional neuraL nEtworks) is proposed in \cite{sankhe2019oracle}. Here, one-dimensional (1D) convolutional layers are adopted to process the incoming complex in-phase and quadrature (IQ) samples. The work shows 99\% identification of WiFi devices (16 Universal Software Radio Peripheral (USRP) X310) with WiFi signal generated using software from GNU Radio \cite{grc}. However, the main drawback of this work is the utilization of transmitter level impairments and the use of synthetically generated WiFi signals. We note here that such artificially introduced impairments mask the subtle hardware imperfections specific to the device which are typically extracted for RF fingerprinting. Further, the applicability and practicality of such artificial perturbations in identifying already deployed non-programmable IoT emitters in the present-day market are slim. Therefore, we state here that in order to accelerate and promote the acceptance of RF fingerprinting solutions in the present-day IoT infrastructure, the fingerprinting systems must utilize the raw IQ samples from the emissions without any additional perturbations in the emitter (transmitter) side. The fewer preprocessing (of received raw samples) steps, the faster the fingerprinting system. 

A massive experimental study was presented in \cite{shawabka2020exposing} to show the effects of the wireless channel and IQ equalization on WiFi and ADS-B device fingerprinting. The work involved fingerprinting of 5000 ADS-B and 5117 WiFi devices to show the impact of the device population on the performance. The authors report the results on a custom software-generated dataset transmitted over the air with up to 20 USRP family of radios under different collection environments and a large-scale dataset provided by DARPA including captures from nearly 10k WiFi and ADS-B devices. The authors adopt different CNN architectures to perform the fingerprinting and report accuracy gains with IQ equalization by up to 23\%. However, even partial equalization of the signals will distort the pure fingerprint of the device.  A long short-term  memory (LSTM) framework to extract the stochastic features from the signal to watermark them into the original signal is proposed in \cite{lstm_watermark} to avoid data injection attacks. In \cite{wifi_rff}, the common Wifi waveform features unique to the IEEE802.11a/g/p such as carrier frequency offset, scrambling seed pattern, sampling frequency offset, and transient ramp-up/down periods are leveraged to distinguish Wifi cards. The authors resorted to software generation of Wifi stack on USRP software-defined radios instead of real-world emissions. An unmanned aerial vehicle (UAV) classification scheme is proposed in \cite{uav_wifi} where the statistical features of the Wifi standard is utilized to facilitate fingerprinting. 


A CNN is used to identify 7 Zigbee devices using the time-domain complex baseband error signal and attains a 92.29\% accuracy in \cite{merchant}. However, the approach involves forward-backward signal filtering with a fourth-order Butterworth filter and a 2 MHz passband whereby the ideal signal is subtracted from the transmission resulting in an error signal. An inter-arrival approach was shown effective for WiFi fingerprinting using a feed-forward neural network in \cite{gtid}. However, these approaches rely on protocol-specific processing and signal modeling in contrast to our proposed approach that leverages unprocessed raw IQ samples.

The works discussed so far rely on either protocol-specific signal processing and/or transmitter-side artificial impairments. However, such approaches rely on strong apriori assumptions on the type of protocol that can be leveraged to trace the device origin. We argue that to ensure ubiquity, transparency, and robustness across capture times, the fingerprinting approach must exploit \textit{raw} unprocessed IQ samples from passive signal reception across diverse wireless protocols. In this work, for the first time, we introduce a novel cross-domain attention mechanism for comprehensive feature extraction by exploiting raw IQ samples collected from a passive listener (radio). Additionally, we demonstrate the performance gain achieved by the adoption of multi-task architecture which jointly fingerprints WiFi and BT devices while also classifying the wireless protocol. While multi-task architecture has not been applied to fingerprinting, the utility of multi-task architecture has been demonstrated in modulation and protocol classification in \cite{AJagannath21ICC,AJagannath22PHYCOM}.

\textit{Attention mechanism} was first introduced in the encoder-decoder architectures of neural machine translation models in natural language processing (NLP)\cite{nmt}. Although attentional learning is prevalent in NLP and computer vision \cite{cv_attn1,cv_attn2}, its adoption in wireless realm has been limited \cite{tf_attn}. To best of our knowledge, attention learning has not  been applied to the RF fingerprinting problem. 
In this work, for the first time in literature, we intend to capture all domains - \textit{spatial, temporal, and time-frequency} - of subtle feature manifestations present in the RF signal emissions to arrive at a comprehensive attentional vector which would be robust across the type of emission, time of capture, and other confounding factors. The novelty and details of the proposed architecture is described in detail in Section \ref{Sec:Arch}

\section{Real World RF Signal Collection}
\subsection{Data collection Framework}
Keeping the practical relevance of RF fingerprinting in mind, we opt for COTS IoT devices instead of generating synthetic standard-compliant waveforms using SDRs using GNU Radio or MATLAB as in majority of the literature. We create a challenging RF dataset by choosing COTS IoT devices. 
The emitters comprise total of 10 devices; 8 identically manufactured Raspberry Pi4Bs and 2 identical laptops from the same manufacturer (Lenovo). 
The Raspberry Pi4B results in a complex device group since they use a Bluetooth and WiFi transceiver combo chip (Cypress CYW43455) whereby parts of the RF circuitry is shared between the Bluetooth and WiFi modules including a shared single dual-band antenna. It is worth noting that the Cypress chip is suitable for and hosted in several industrial and smart home IoT devices. 
Similarly, the Lenovo laptops possess a Intel Dual Band Wireless-AC 7260 combo chip supporting WiFi and Bluetooth. The chip has two antenna ports - one reserved for WiFi while the other is shared between WiFi and Bluetooth.

\begin{figure}[h!]
\centering
\includegraphics[width=.88\columnwidth]{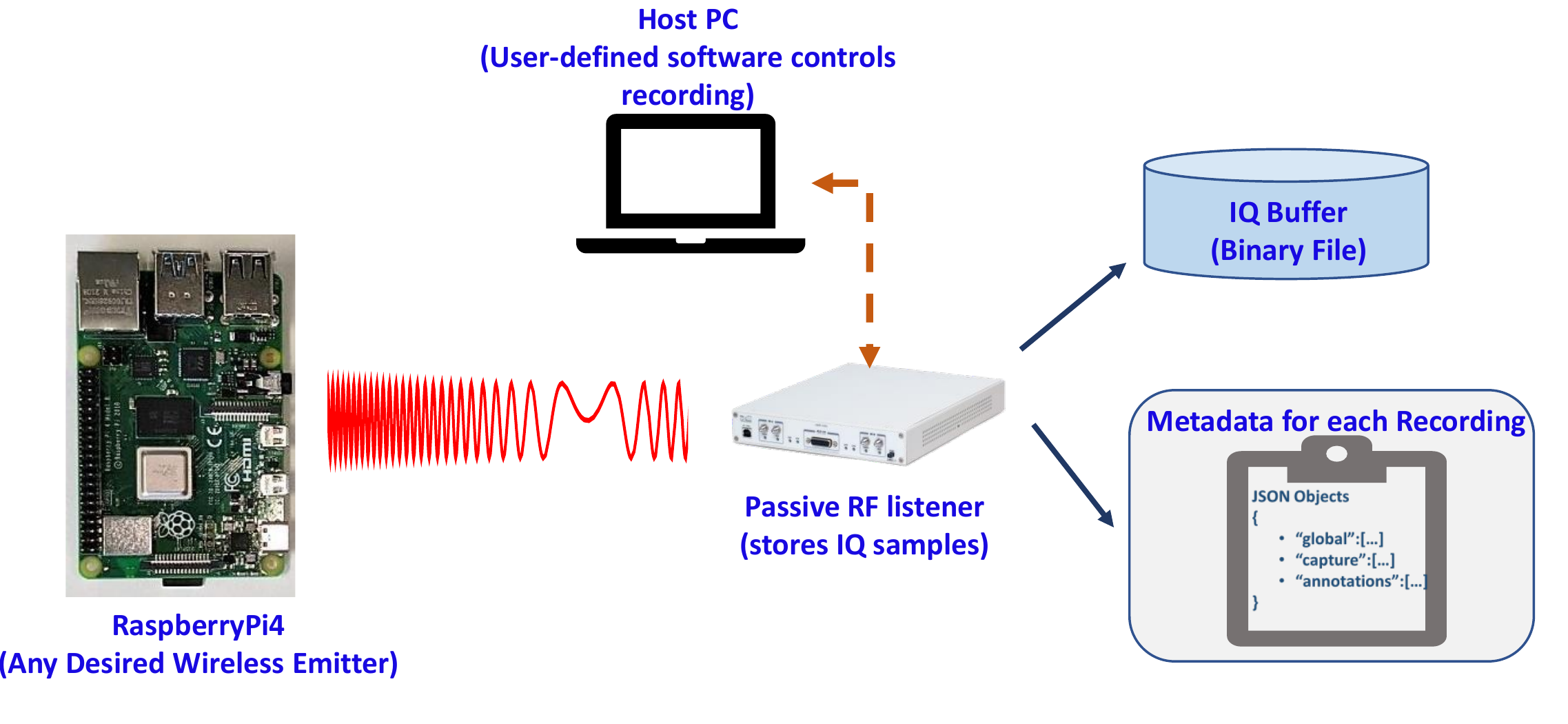} 
\caption{Experimental data collection framework}
\label{fig:data}
\end{figure}

In this section, we elaborate the data collection framework to enable IoT device fingerprinting. 
The IoT emitter refers to any IoT device that we desire to fingerprint from the device-specific minuscule imperfections (signature) embedded into the emitted waveform. 
The receiving radio scans a wide frequency band within which the emitter transmits. The radio we use for spectrum sensing and capture is a USRPX300 outfitted with a UBX160 daughtercard and VERT2450 antenna for scanning the 2.4 GHz ISM band. The spectrum sensing radio is centered at 2.414 GHz and receives samples at the rate of 66.67 MS/s. The captures are saved as complex IQ samples in a binary file with each recording sized at 40 MS. For each capture, an associated metadata compliant with the latest SigMF specifications and additional field extensions for usability are recorded. The RF fingerprinting testbed framework is shown in Fig.\ref{fig:data} with the emitters and passive listener located in the indoor laboratory amidst unavoidable nearby emitters, human mobility, electronic hums, other indoor obstacles forming a rich indoor multipath environment. 
Hence, we consider this as a \textit{real-world in-the-wild} data collection environment.
WiFi and Bluetooth are two ubiquitous wireless standards that are present in most of the IoT devices and hence most commonly found in the RF spectrum. Accordingly, instead of choosing either one of them or choosing any openly available simulated and/or synthetically generated datasets, we select both wireless standards for real-world data capture from IoT device emissions.

\textbf{\textit{IEEE802.11g WiFi emissions}}: The aforementioned IoT devices are the source of the WiFi emissions which operates with IEEE802.11g at the WiFi channel 8 centered at 2.447 GHz. 
We resort to unequalized and unfiltered raw IQ samples to preserve the transmitter-specific features present in the captured waveform to its entirety. 

\begin{figure}
\centering
\includegraphics[width=.99\columnwidth]{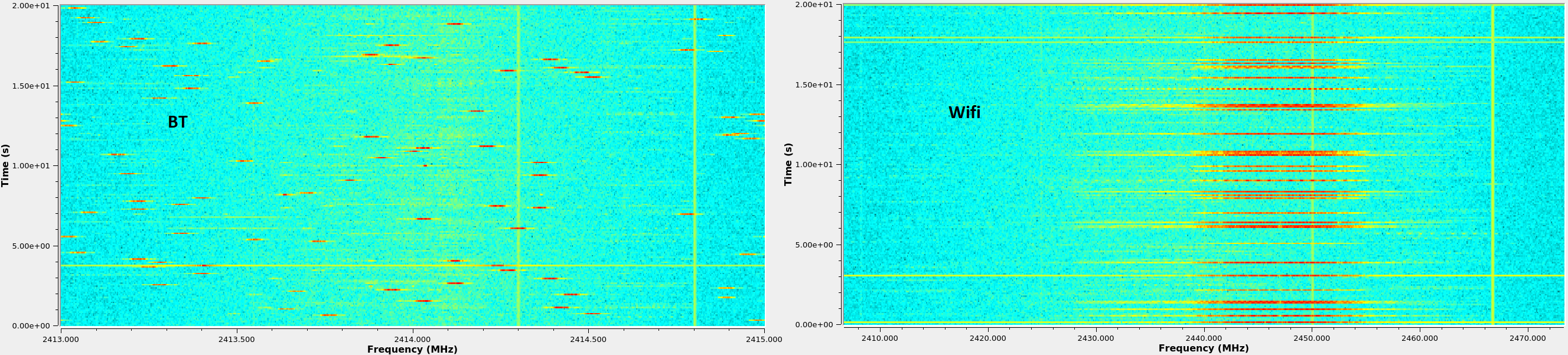}
\caption{BT (left) and WiFi (right) signals from same device.}
\label{fig:bt_wifi}
\vspace{-0.5 cm}
\end{figure}

\textbf{\textit{Bluetooth emissions}}: BT signal due to its frequency hopping nature presents a challenging waveform category for RF fingerprinting. The LO on the combo chip provides fast frequency hopping (1600 hops/second) over the supported BT channels. 
The emission of Bluetooth and WiFi from same device is demonstrated as a time-frequency waterfall spectrogram in Fig.\ref{fig:bt_wifi}. The time-frequency view of the spectrum clearly shows the frequency-hopping nature of the Bluetooth waveform which aggravates the fingerprinting problem.

\section{Cross-domain attentional architecture}
\label{Sec:Arch}

\textit{Attentional learning or attention mechanism} 
is the neural network’s attempt at mimicking the brain by selectively focusing on few relevant areas of a given input. 
Visual attention has demonstrated immense benefit in structural prediction tasks such as image/video captioning, visual quizzing, etc \cite{cv_attn1,cv_attn2}. 
Recurrent neural network (RNN) on the other hand, extracts the periodicity or the \textit{temporal} pattern present in the input. However, they do not perform well with long input sequences since RNN units suffer from long-range dependency issues due to vanishing/exploding gradients. Even LSTM cells tend to become forgetful in learning long sentences. This is where the introduction of the attention mechanism resolved this major drawback in the NLP domain. 
Luong et al. \cite{luong-etal} classify attention into two categories; global and local depending on where the attention is applied. In the context of NLP, the goal of attention is to derive a context vector that captures relevant source-side information to aid in the target word prediction. Specifically, the authors resort to a simple concatenation layer to concatenate the source-side context vector with the target hidden state to generate an attentional hidden state. Another impactful work was by Vaswani et al. \cite{vaswani} where a transformer model is built upon self-attention without using sequence aligned recurrent architectures. Self-attention also called intra-attention refers to relating different positions of a single sequence in order to compute a representation of the same sequence. With attention, each element of the context vector is given relative importance by employing weights allowing the network to learn the most significant portions of sentences.


Inspired from these advances, we propose novel attentional architecture - Cross-domain attentional model (xDom) (shown in Fig.\ref{fig:xdom}) for RF fingerprinting. xDom is a multi-task architecture in its construction implying it can perform joint multiple predictions with a single neural network model for a given input. The multi-task predictions can be any related tasks that can be derived from the same input presentation. Here, xDom is designed to perform RF fingerprinting and wireless protocol classification. The architecture ingests 1024 complex IQ samples in a 2D tensor format (arranged as rows of the 2D tensor). xDom adopts three domains of feature extraction:
\begin{figure}[h!]
\centering
\includegraphics[width=.99\columnwidth]{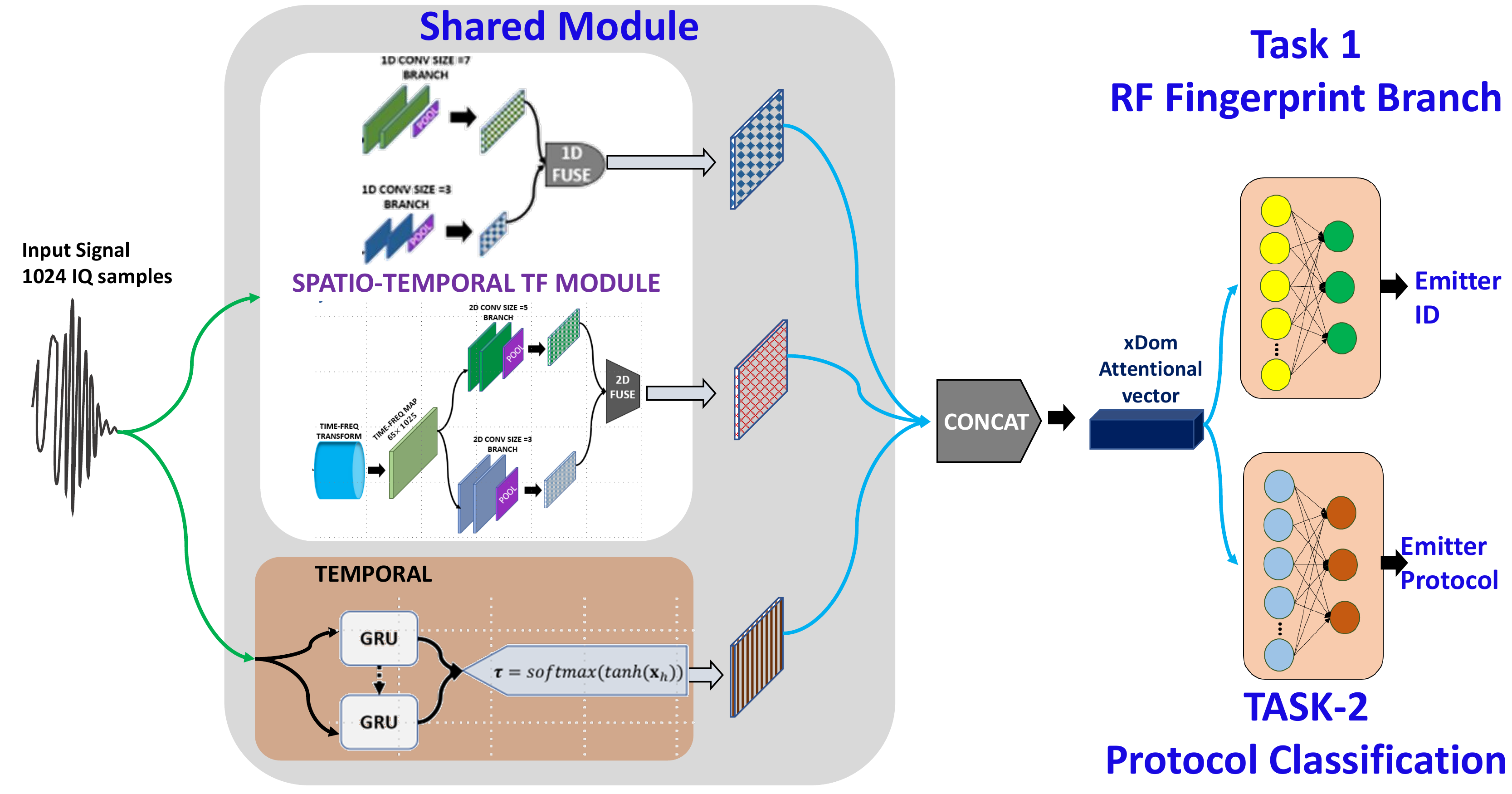} 
\caption{Proposed xDom multi-task architecture}
\label{fig:xdom}
\end{figure}
\begin{enumerate}
    \item \textbf{\textit{Spatio-temporal features}}: xDom adopts a spatial filtering module which is a bank of 1D and 2D convolutional layers. There are two parallel 1D-convolutional (conv1D) layers to process the ingested 1024 complex IQ samples and another set of parallel 2D-convolutional (conv2D) layers to spatially filter the input TF map. The conv1D filter banks process the IQ input as a 2-channel input to extract the local temporal correlations from the two input channels. The conv2D filter bank on the other hand spans across the 2D TF map of size $65\times1025$ to extract the prominent spatial TF features. The magnitude and phase of the TF map are separately processed in each of the conv2D branches for a comprehensive representation.
    \item \textbf{\textit{Temporal pattern}}: The input IQ tensor is fed through a temporal module to extract any specific temporal patterns arising from the nature of the waveform and/or the hardware imperfections. The temporal module comprises a two-layer gated recurrent unit (GRU) with 132 hidden units. We resort to GRU instead of LSTM as they are comparatively easy (faster) to train and utilize only lesser memory. The output ($\mathbf{x}_o$) from the temporal module is concatenated to the hidden state ($\mathbf{h}$) resulting in a concatenated vector ($\mathbf{x}_h$).
    \begin{equation}
        \mathbf{x}_h = vec\left(\mathbf{x}_o:\mathbf{h} \right)
    \end{equation}
    where $vec\left(\cdot\right)$ and $:$ are the vectorization and concatenation operator. Specifically, xDom adopts a many-to-1 mapping GRU such that it outputs a $1\times132$ vector instead of a $1024\times132$ matrix. The hidden state is of dimension $1\times132$ resulting in a $1\times264$ concatenated vector.
    \item \textbf{\textit{Time-Frequency mapping}}: xDom utilizes a runtime short-time Fourier transform (STFT) block which maps the input IQ to 2D TF map $\mathbf{x}_{TF}$. The STFT performs a 128-point FFT operation to produce a $65\times1025$ output TF map. The TF map is split into its component magnitude and phase representations to pass it through the conv2D filter bank.
\end{enumerate}

Following the temporal module is a single layer linear feed-forward neural network with hyperbolic tangent (tanh) activation and a softmax mapping which maps the concatenated temporal pattern vector $\mathbf{x}_h$ into an attentional scoring vector ($\tau$) as in
\begin{equation}
    \mathbf{\tau} = softmax\Big(tanh\big( \mathbf{x}_h \big) \Big)
\end{equation}
Here, the softmax function yields the output score from the feed-forward neural network output vector which essentially is the temporal scoring. Intuitively, this scoring accounts for the saliency captured by the temporal feature vector ($\mathbf{x}_h$). Now, the cross-domain attentional vector ($\mathbf{a}_{\text{xdom}}$) is derived by the following operation,
\begin{equation}
    \mathbf{a}_{\text{xdom}} = vec\Big( \mathbf{x}_1^{IQ}:\mathbf{x}_2^{IQ}:vec\big( \mathbf{x}_3^{phase}:\mathbf{x}_4^{mag}\big):\mathbf{\tau} \Big)
\end{equation}

This comprehensive attentional feature vector captures the essence of the different perturbations present in the RF emissions and can now be leveraged to perform the relevant fingerprinting classification. The attentional feature vector $\mathbf{a}_{\text{xdom}}$ is split into two classifier branches which are simple feed-forward neural network layers where the final output layer performs softmax classification. Here, the two classifier modules perform fingerprinting and wireless protocol classification.

\section{Over-the-air Experimental Evaluation}
\subsection{Evaluation Setups}

We evaluate the proposed xDom against a well-known CNN model for RF fingerprinting \cite{shawabka2020exposing} under real-world evaluation setups. We select the Baseline CNN model in \cite{shawabka2020exposing} and for ease refer to it as AlshBaseline. The proposed xDom architecture supports single and multi-task forms. However, in order to be fair to the benchmark architecture which was originally designed for singular task - fingerprinting (identifying device origin), we resort to showing the evaluation results in the single as well as multi-task settings. Both xDom and AlshBaseline were trained by splitting the dataset into 70-15-15 training, validation, and testing sets. The models were implemented in the PyTorch framework and trained with stochastic gradient descent (SGD) optimizer with a learning rate of 0.1  and momentum of 0.9 for 150 epochs. 

We test the models under single-task and multi-task settings to show the performance gain achieved by xDom architecture. Under single-task settings, the proposed model would only have a single classifier branch which performs RF fingerprinting (emitter classification). Similarly, with regards to the benchmark architecture, we set up two softmax classifier branches to perform the multi-task emitter and protocol classifications. Under single-task setting, the BT and WiFi signals are separately used in the dataset whereas for multi-task classification, the dataset contains equal proportion of BT and WiFi signals.
\begin{figure*}[h!]
     \centering
     \begin{subfigure}[b]{0.3\textwidth}
         \centering
         \includegraphics[width=\textwidth,height=3.7cm]{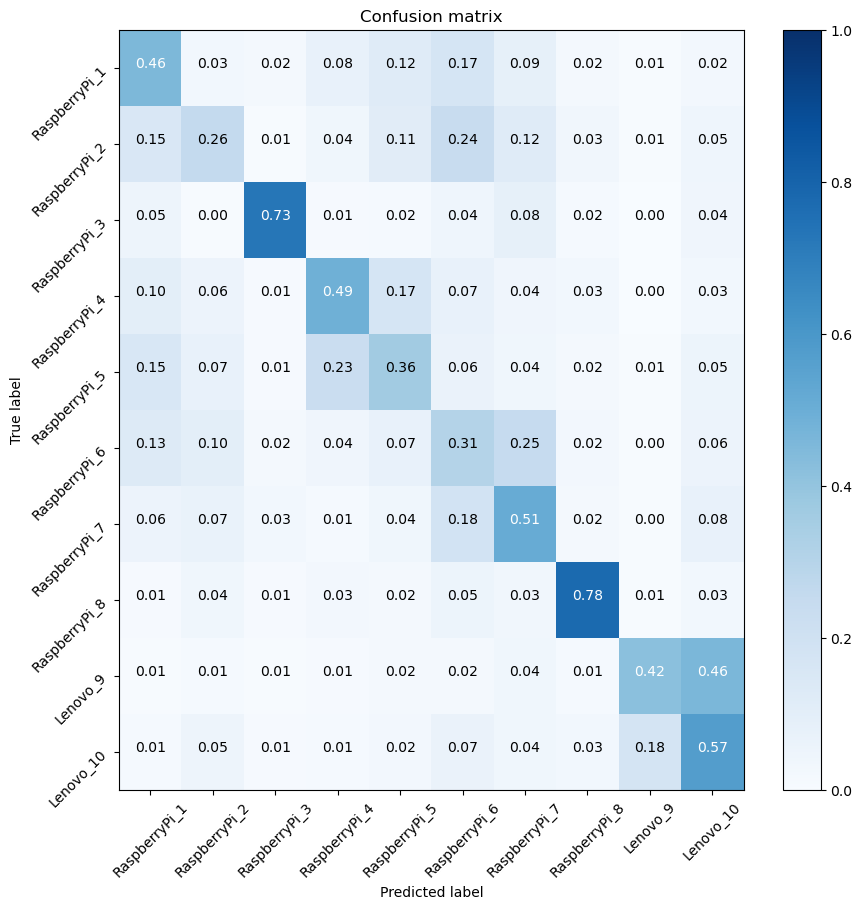}
         \caption{STL: BT fingerprinting}
         \label{fig:ttsd_xdom_bt}
     \end{subfigure}
     \hfill
     \begin{subfigure}[b]{0.3\textwidth}
         \centering
         \includegraphics[width=\textwidth, height=3.7cm]{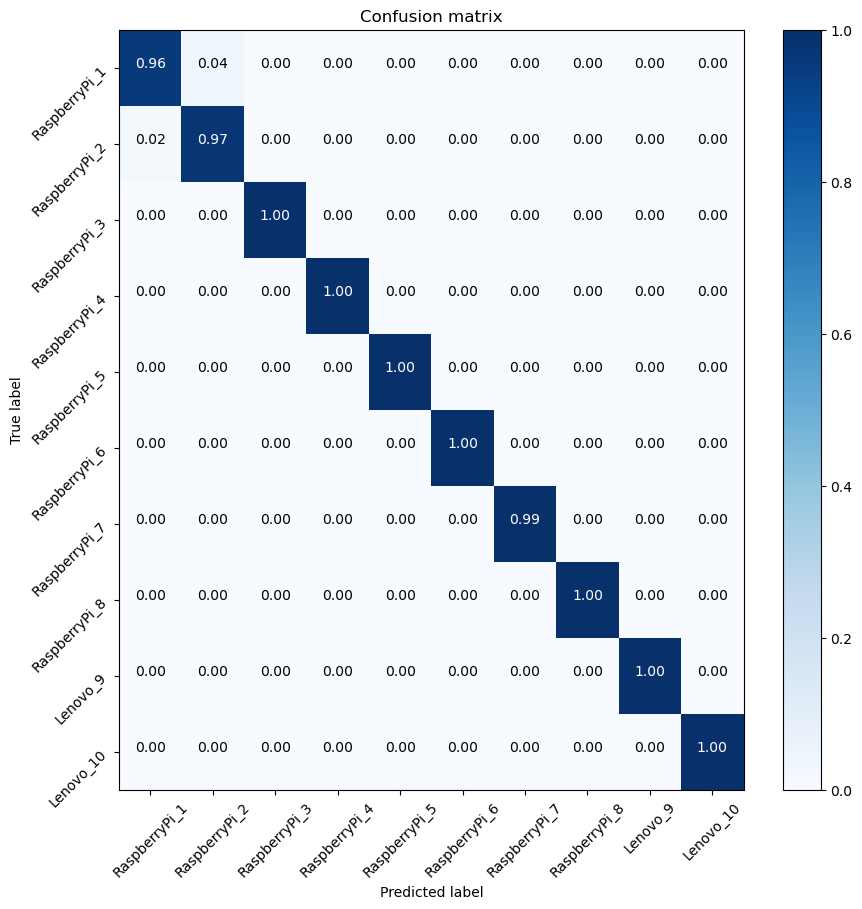}
         \caption{STL: WiFi fingerprinting}
         \label{fig:ttsd_xdom_wifi}
     \end{subfigure}
     \hfill
         \begin{subfigure}[b]{0.3\textwidth}
         \centering
         \includegraphics[width=\textwidth, height=3.7cm]{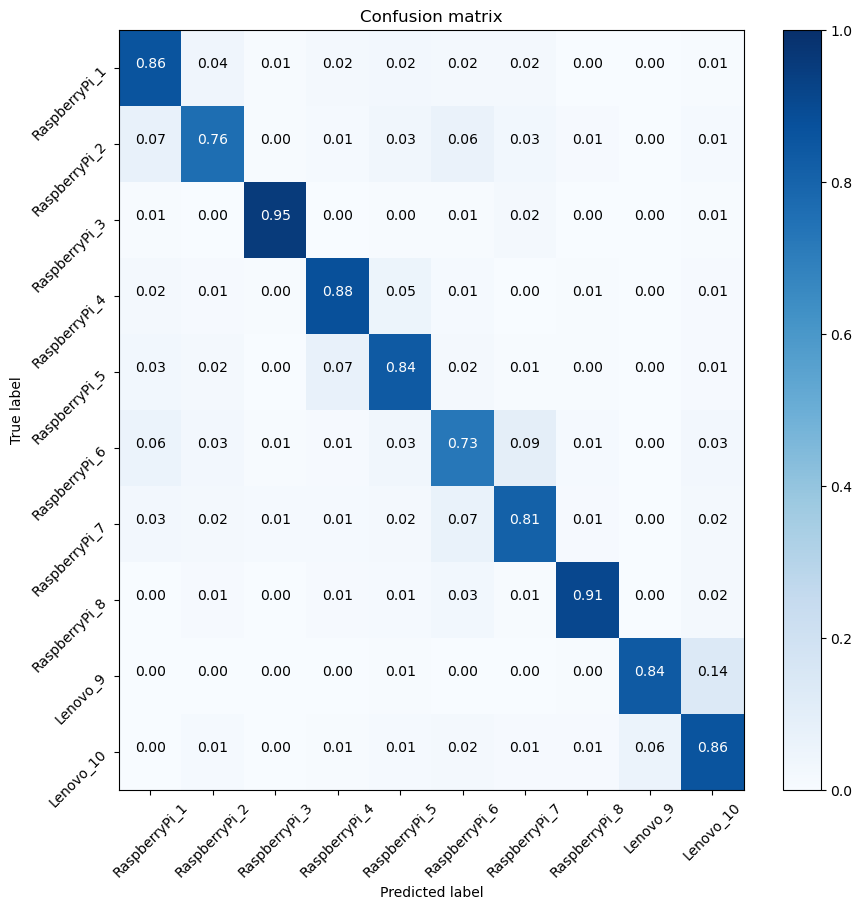}
         \caption{MTL: joint BT and WiFi fingerprinting}
         \label{fig:ttsd_xdom_mtl}
     \end{subfigure}
        \caption{TTSD: Proposed xDom architecture under single-task (BT and WiFi) and multi-task (WiFi $+$ BT) settings}
        \label{fig:ttsd_xdom}
\end{figure*}

\begin{figure*}[h!]
     \centering
     \begin{subfigure}[b]{0.3\textwidth}
         \centering
         \includegraphics[width=\textwidth, height=3.7cm]{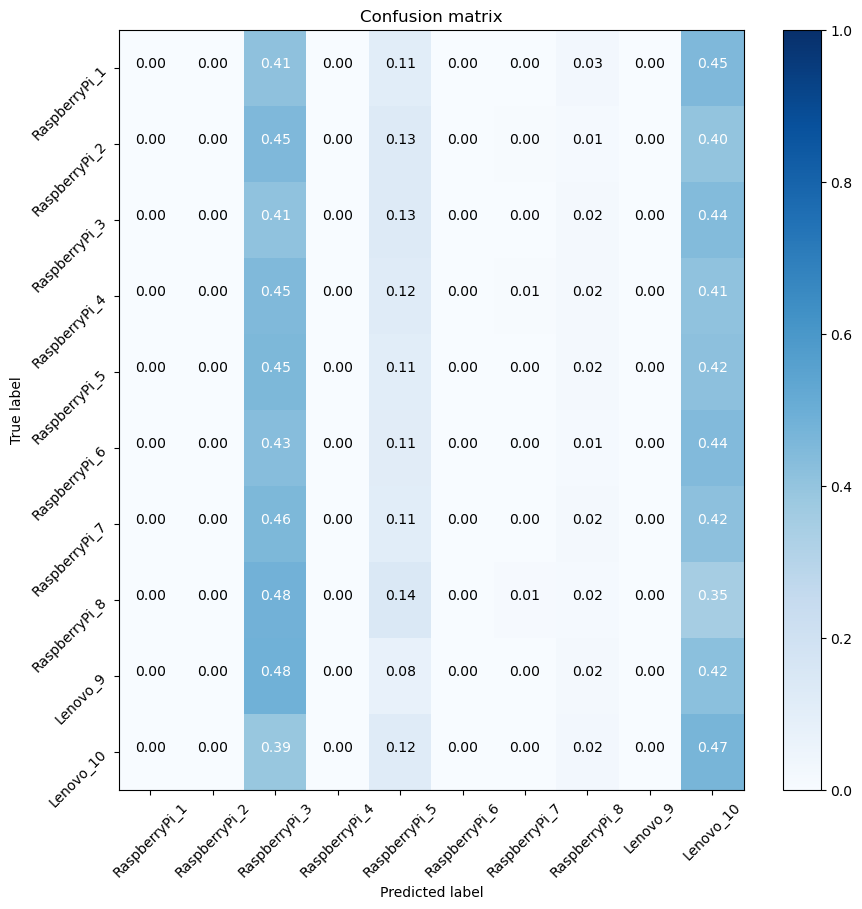}
         \caption{STL: BT fingerprinting}
         \label{fig:ttsd_alsh_bt}
     \end{subfigure}
     \hfill
     \begin{subfigure}[b]{0.3\textwidth}
         \centering
         \includegraphics[width=\textwidth, height=3.7cm]{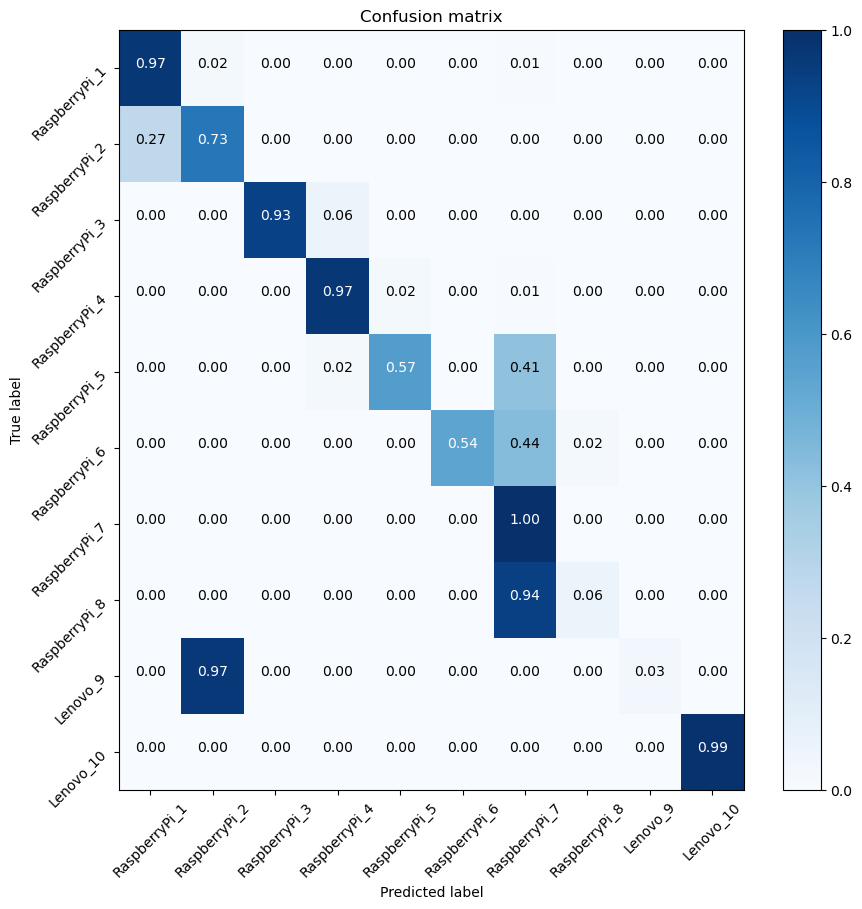}
         \caption{STL: WiFi fingerprinting}
         \label{fig:ttsd_alsh_wifi}
     \end{subfigure}
     \hfill
         \begin{subfigure}[b]{0.3\textwidth}
         \centering
         \includegraphics[width=\textwidth, height=3.7cm]{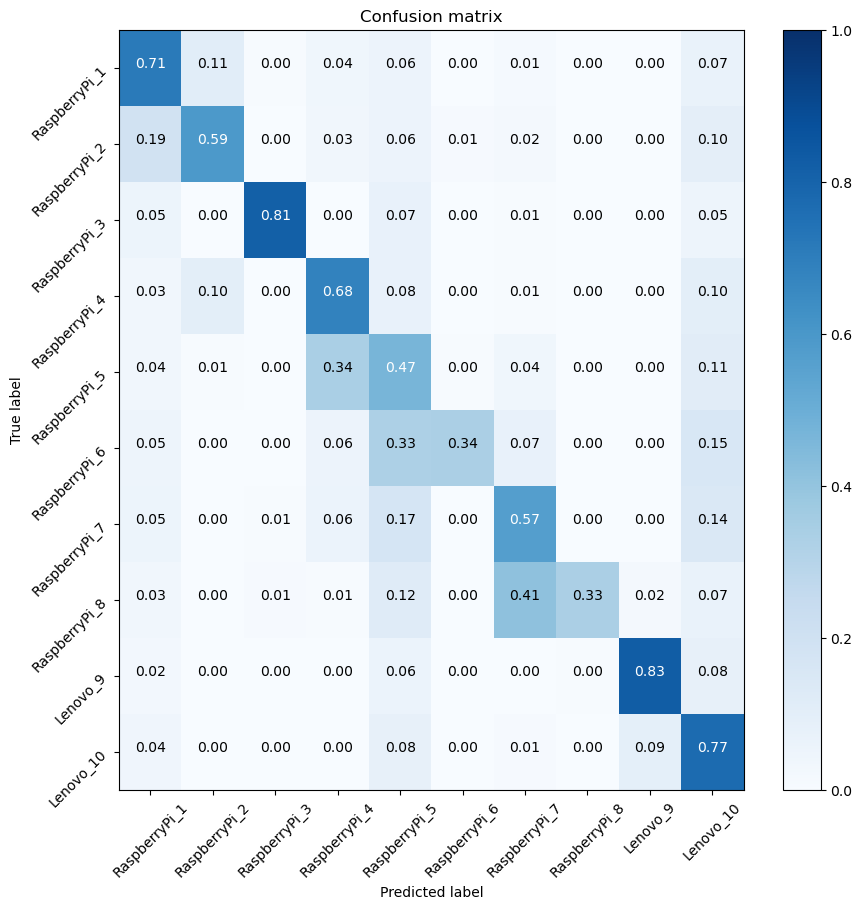}
         \caption{MTL: joint BT and WiFi fingerprinting}
         \label{fig:ttsd_alsh_mtl}
     \end{subfigure}
        \caption{TTSD: Benchmark AlshBaseline architecture under single-task (BT and WiFi) and multi-task (WiFi $+$ BT) settings}
        \label{fig:ttsd_alsh}
\end{figure*}

The architectures will be evaluated under two broad scenarios: Train and Test Same Day (TTSD) and Train and Test Mixed Days (TTMD). In the TTSD setup, the models will be trained, validated, and tested with the dataset captured in the same time frame. On the other hand, TTMD would have dataset collected over a longer time frame (spanning months) such that it would capture the environmental uncertainties over multiple days and hence possess a richer and heterogeneous data distribution.


\subsection{Evaluation Results}

\textbf{xDom Performance:} Figure \ref{fig:ttsd_xdom} demonstrates the top-1 fingerprinting accuracy and false alarm rates of the proposed xDom architecture under the single and multi-task settings. The single-task BT fingerprinting achieves a highest of 78\% accuracy and a lowest of 26\% at low false alarm rates as shown in Fig.\ref{fig:ttsd_xdom_bt}. This can be resorted to the challenging frequency-hopping nature of the BT emissions. On the other hand, the WiFi fingerprinting achieves a 100\% accuracy with a lowest of 96\% and negligible false alarms as seen in Fig.\ref{fig:ttsd_xdom_wifi}. These single-task performances are significantly improved under the multi-task setting (Fig.\ref{fig:ttsd_xdom_mtl}) which performs joint WiFi and BT fingerprinting and protocol classification. The average top-1 fingerprinting accuracy under the multi-task setting is 84.3\%. We benchmark these performances against the AlshBaseline. 

\textbf{AlshBaseline Performance:} Figure \ref{fig:ttsd_alsh_bt} shows that this architecture fails to perform any usable BT fingerprinting. Although, WiFi fingerprinting (Fig.\ref{fig:ttsd_alsh_wifi}) of shows some improvement over BT (Fig.\ref{fig:ttsd_alsh_bt}), it has very high false alarm rates. Even in the case of the AlshBaseline Fig.\ref{fig:ttsd_alsh_mtl}, it can be observed that the performance of the multi-task version improves with higher fingerprinting accuracy.

\begin{figure*}
     \centering
     \begin{subfigure}[b]{0.3\textwidth}
         \centering
         \includegraphics[width=\textwidth, , height=3.7cm]{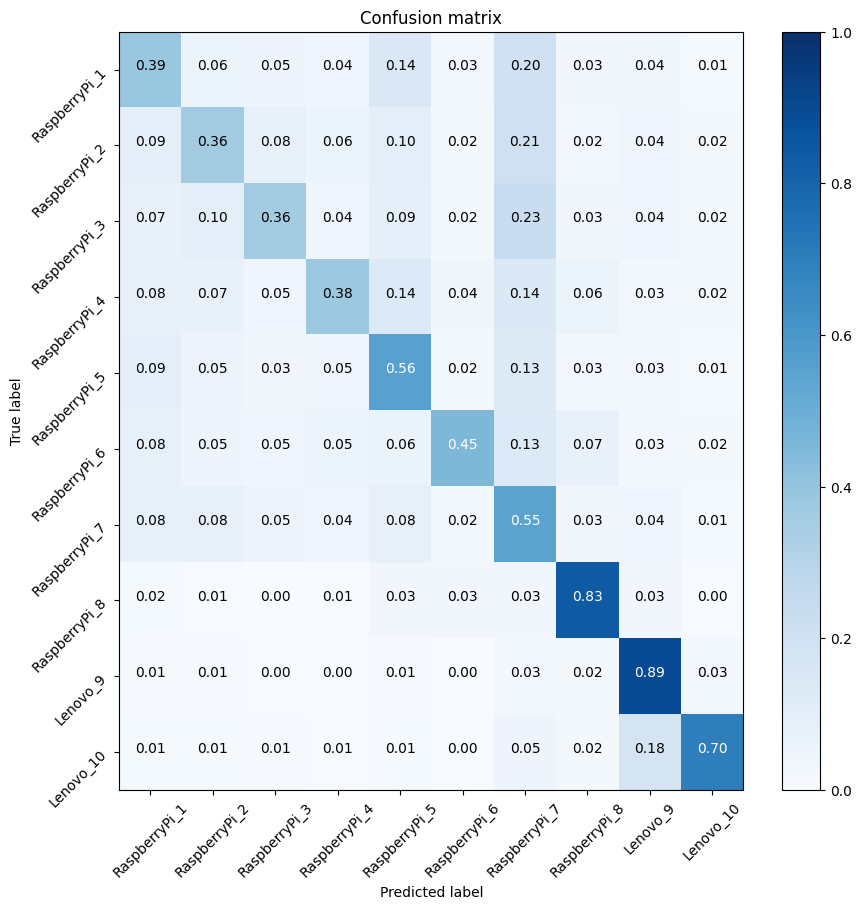}
         \caption{STL: BT fingerprinting}
         \label{fig:ttmd_xdom_bt}
     \end{subfigure}
     \hfill
     \begin{subfigure}[b]{0.3\textwidth}
         \centering
         \includegraphics[width=\textwidth, height=3.7cm]{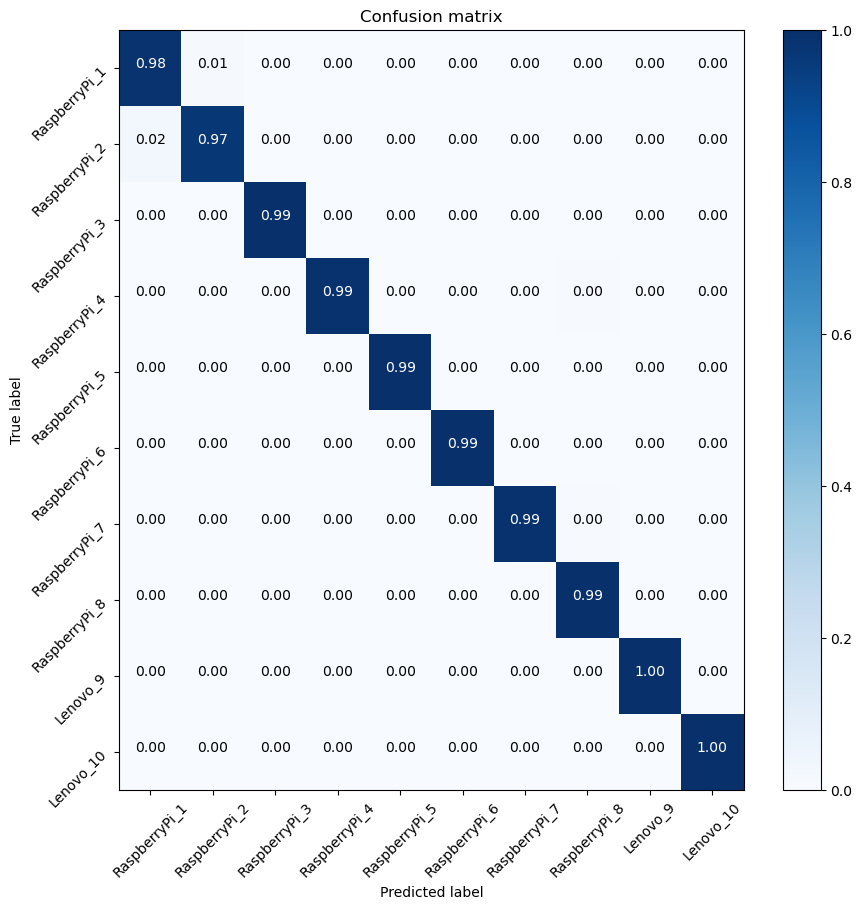}
         \caption{STL: WiFi fingerprinting}
         \label{fig:ttmd_xdom_wifi}
     \end{subfigure}
     \hfill
         \begin{subfigure}[b]{0.3\textwidth}
         \centering
         \includegraphics[width=\textwidth, height=3.7cm]{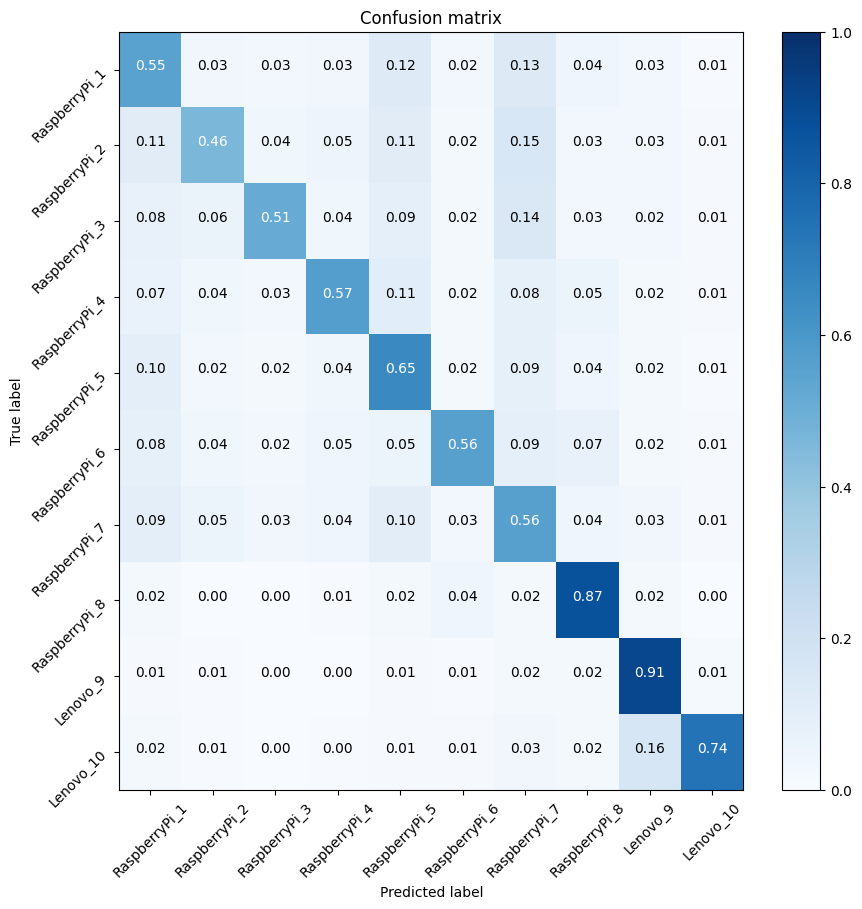}
         \caption{MTL: joint BT and WiFi fingerprinting}
         \label{fig:ttmd_xdom_mtl}
     \end{subfigure}
        \caption{TTMD: Proposed xDom architecture under single-task (BT and WiFi) and multi-task (WiFi $+$ BT) settings}
        \label{fig:ttmd_xdom}
        \vspace{-.4 cm}
\end{figure*}

\begin{figure*}
     \centering
     \begin{subfigure}[b]{0.3\textwidth}
         \centering
         \includegraphics[width=\textwidth, height=3.5cm]{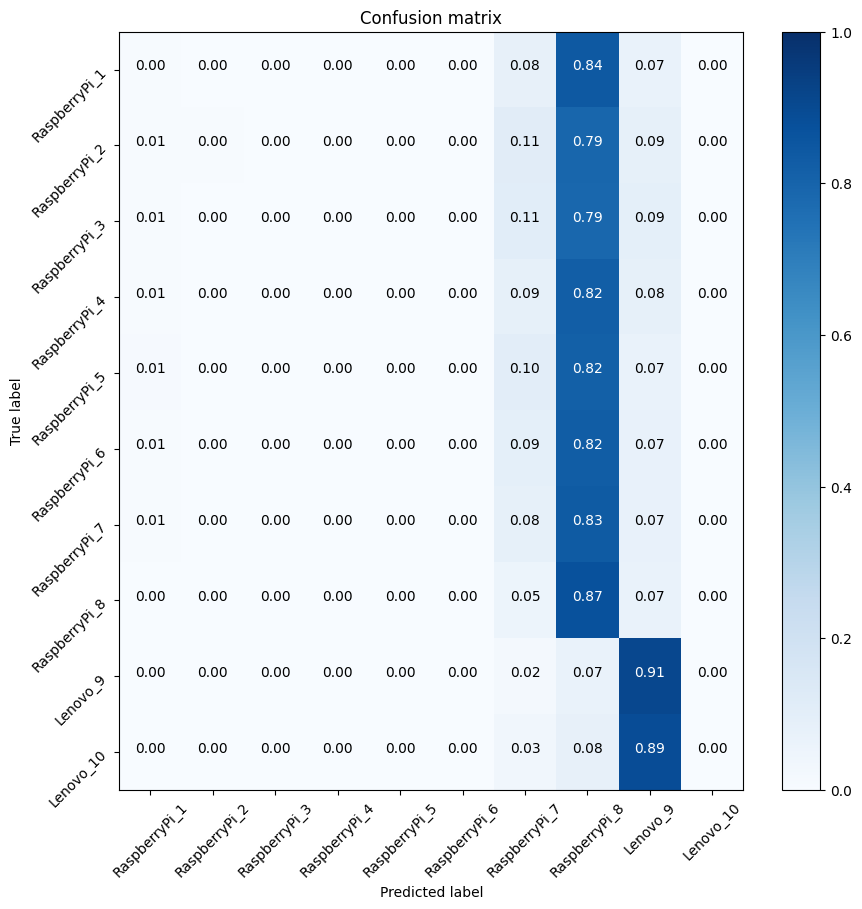}
         \caption{STL: BT fingerprinting}
         \label{fig:ttmd_alsh_bt}
     \end{subfigure}
     \hfill
     \begin{subfigure}[b]{0.3\textwidth}
         \centering
         \includegraphics[width=\textwidth, height=3.5cm]{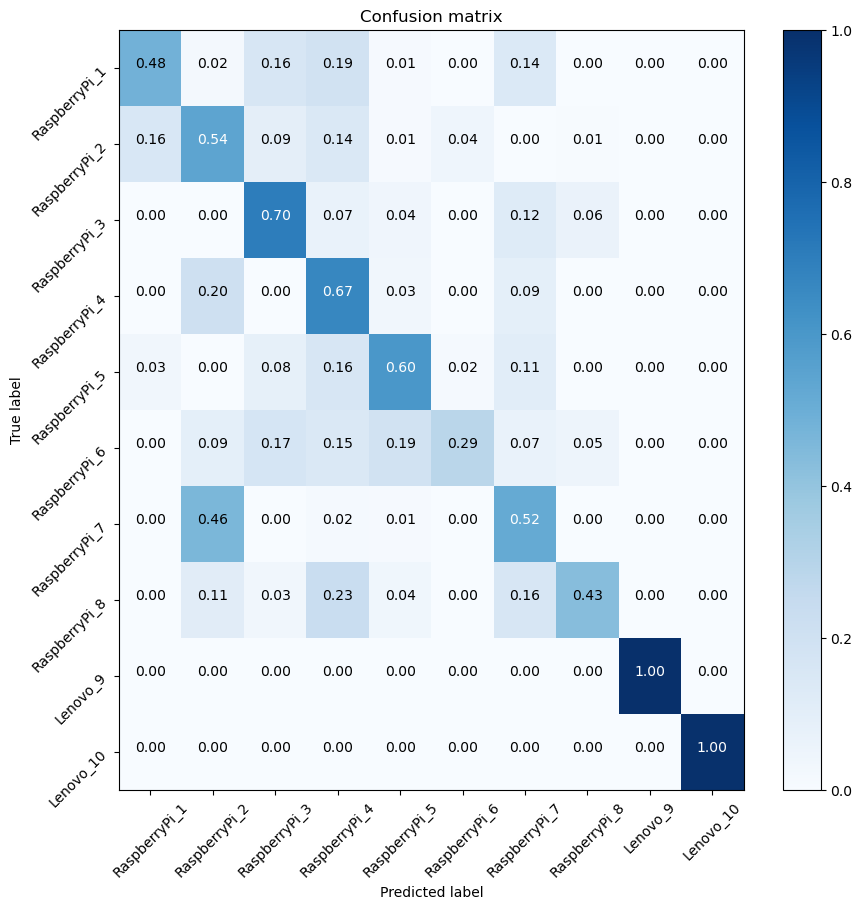}
         \caption{STL: WiFi fingerprinting}
         \label{fig:ttmd_alsh_wifi}
     \end{subfigure}
     \hfill
         \begin{subfigure}[b]{0.3\textwidth}
         \centering
         \includegraphics[width=\textwidth, height=3.5cm]{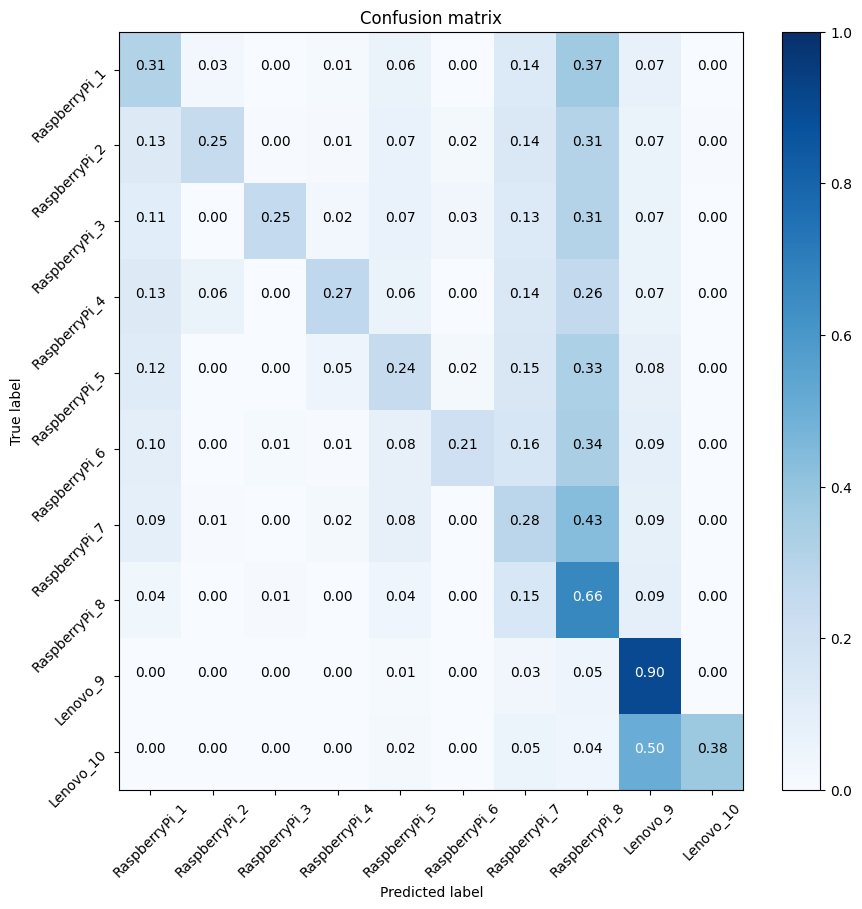}
         \caption{MTL: joint BT and WiFi fingerprinting}
         \label{fig:ttmd_alsh_mtl}
     \end{subfigure}
        \caption{TTMD: Benchmark AlshBaseline architecture under single-task (BT and WiFi) and multi-task (WiFi $+$ BT) settings}
        \label{fig:ttmd_alsh}
       \vspace{-.2cm}
\end{figure*}

The same performance trend can be seen with the richer dataset under the TTMD setup as shown in Fig.\ref{fig:ttmd_xdom} and Fig.\ref{fig:ttmd_alsh}. BT fingerprinting is equally poor as with TTSD with the AlshBaseline architecture (Fig.\ref{fig:ttmd_alsh_bt}). However, it is evident from Fig.\ref{fig:ttmd_alsh_wifi} that by presenting a richer diverse dataset under the TTMD setup, we note significant performance improvement with the WiFi fingerprinting showing fewer false alarms and improved fingerprinting accuracy in contrast to Fig.\ref{fig:ttsd_alsh_wifi}. 

\textbf{xDom vs AlshBaseline:} As summarized in Table \ref{tab:stl_eval}, the proposed xDom attains 45.95\% and 59.3\% higher accuracy in WiFi fingerprinting under the TTSD and TTMD settings respectively in contrast to the AlshBaseline. Similarly, BT fingerprinting performs 4.91$\times$ and 2.92$\times$ higher than the AlshBaseline under TTSD and TTMD setups respectively. Under the multi-task setting in Table \ref{tab:mtl_eval}, xDom scores a fingerprinting accuracy of 84.3\% and 63.8\% with the TTSD and TTMD setups respectively in contrast to just 60.1\% (TTSD) and 37.6\% (TTMD) attained with the AlshBaseline. 
These significant performance gains over AlshBaseline exhibit the effectiveness of adopting attentional learning with the proposed xDom architecture which analyzes and extracts features from diverse dataset domains.

\begin{table}
\centering
\caption{Performance of xDom under single-task setting. \label{tab:stl_eval}}
\begin{tcolorbox}[width=3.5in,tab2,tabularx={p{1.3 cm}|p{1.5 cm}|p{2.2 cm}p{2.2 cm}}]
\textbf{Scenario} & \textbf{Waveform} & \multicolumn{2}{c}{\textbf{Top-1 Fingerprinting Accuracy}} \\
& & xDom &AlshBaseline\\
\hline
TTSD &WiFi &\textbf{0.991} &0.679\\
& BT &\textbf{0.487} &0.099\\\hline
TTMD &WiFi & \textbf{0.991} &0.622\\
& BT & \textbf{0.546}
&0.187
\\\hline
\end{tcolorbox}
\vspace{-.8 cm}
\end{table}

The multi-task architecture evidently presents improved joint WiFi and BT fingerprinting accuracy under both evaluation setups suggesting that the model learns a more intuitive feature presentation by incorporating diverse protocol emissions allowing it to distinctly identify the innate hardware perturbations that contribute to the device fingerprint.  
While we have shown the effectiveness of xDom in single-task and multi-task setting, the advantage gained by adopting a multi-task architecture will greatly benefit the next generation of wireless systems comprising devices with heterogeneous wireless standards.

\begin{table*}[h!]
    \centering
    \caption{Performance of xDom under multi-task setting. \label{tab:mtl_eval}}
\begin{tcolorbox}[width=6.2in,tab2,tabularx={p{1.5 cm}|p{3.1 cm}p{3.1 cm}|p{3.1 cm}p{3.1 cm}}]
\textbf{Scenario} & \multicolumn{2}{|c|}{\textbf{Top-1 Protocol Identification Accuracy}} & \multicolumn{2}{c}{\textbf{Top-1 fingerprinting Accuracy}} \\
& xDom &AlshBaseline & xDom &AlshBaseline\\
\hline
TTSD & \textbf{1.0} &0.96 &\textbf{0.843} &0.601\\\hline
TTMD &\textbf{1.0}  &0.945
&\textbf{0.638}
&0.376
\\\hline
\end{tcolorbox}
\vspace{-.5cm}
\end{table*}

\section{Conclusion and Future Work}
We presented xDom - a first-of-its-kind cross-domain attentional multi-task architecture for enhancing RF fingerprinting performance. The novel attention mechanism which encompasses all domains of the signal manifestations - Spatio-temporal, temporal, and time-frequency - was elaborated in-depth. The gain in fingerprinting accuracy with the attention mechanism was experimentally demonstrated by comparing against a well-known CNN-based fingerprinting model. Furthermore, the effectiveness of employing the multi-task model was depicted by conducting evaluations under single-task settings. Finally, the analyses were carried out under same (TTSD) and different time-frame settings (TTMD) to study the impact of variability in wireless channels across different time spans. In the future, we plan on improving the single-task BT fingerprinting by incorporating robustness enhancing techniques such as longer-duration IQ samples to overcome the frequency hopping challenge. Additionally, we also plan on curating the vast dataset of real-world IoT emissions for open release to benefit the wider research community. The future work will also incorporate additional evaluations in challenging scenarios.
\bibliography{bibfile.bib}

\begin{thebibliography}{10}
\providecommand{\url}[1]{#1}
\csname url@samestyle\endcsname
\providecommand{\newblock}{\relax}
\providecommand{\bibinfo}[2]{#2}
\providecommand{\BIBentrySTDinterwordspacing}{\spaceskip=0pt\relax}
\providecommand{\BIBentryALTinterwordstretchfactor}{4}
\providecommand{\BIBentryALTinterwordspacing}{\spaceskip=\fontdimen2\font plus
\BIBentryALTinterwordstretchfactor\fontdimen3\font minus
  \fontdimen4\font\relax}
\providecommand{\BIBforeignlanguage}[2]{{%
\expandafter\ifx\csname l@#1\endcsname\relax
\typeout{** WARNING: IEEEtran.bst: No hyphenation pattern has been}%
\typeout{** loaded for the language `#1'. Using the pattern for}%
\typeout{** the default language instead.}%
\else
\language=\csname l@#1\endcsname
\fi
#2}}
\providecommand{\BIBdecl}{\relax}
\BIBdecl

\bibitem{qualcomm}
A.~Davidson, ``{Pushing the limits of Wi-Fi performance with Wi-Fi 7},''
  OnQBlog, 14 February 2022
  \url{https://www.qualcomm.com/news/onq/2022/02/14/pushing-limits-wi-fi-performance-wi-fi-7}.

\bibitem{stat}
F.~Laricchia, ``{Global Bluetooth device shipments 2015-2026},'' Statista, 31
  March 2022
  \url{https://www.statista.com/statistics/1220933/global-bluetooth-device-shipment-forecast/#:~:text=Annual\%20Bluetooth\%20device\%20shipments\%20worldwide\%20stood\%20at\%204.7\%20billion\%20units\%20in\%202021.}

\bibitem{RFfingerChallenges}
Q.~{Xu}, R.~{Zheng}, W.~{Saad}, and Z.~{Han}, ``Device fingerprinting in
  wireless networks: Challenges and opportunities,'' \emph{IEEE Communications
  Surveys Tutorials}, vol.~18, no.~1, pp. 94--104, 2016.

\bibitem{Ajagannath2022ComST2022}
A.~Jagannath, J.~Jagannath, and P.~S. P.~V. Kumar, ``A comprehensive survey on
  radio frequency (rf) fingerprinting: Traditional approaches, deep learning,
  and open challenges,'' \emph{arXiv:2201.00680}, 2022.

\bibitem{JagannathAdHoc2019}
J.~Jagannath, N.~Polosky, A.~Jagannath, F.~Restuccia, and T.~Melodia, ``Machine
  learning for wireless communicationsin the internet of things: A
  comprehensive survey,'' \emph{Ad Hoc Networks (Elseier)}, vol.~93, p. 101913,
  2019.

\bibitem{AJagannath22PHYCOM}
A.~Jagannath and J.~Jagannath, ``{Multi-task Learning Approach for Modulation
  and Wireless Signal Classification for 5G and Beyond: Edge Deployment via
  Model Compression},'' \emph{Physical Communications (Elsevier)}, vol.~54, p.
  101793, 2022.

\bibitem{sankhe2019oracle}
K.~Sankhe, M.~Belgiovine, F.~Zhou, S.~Riyaz, S.~Ioannidis, and K.~Chowdhury,
  ``Oracle: Optimized radio classification through convolutional neural
  networks,'' in \emph{IEEE INFOCOM 2019-IEEE Conference on Computer
  Communications}.\hskip 1em plus 0.5em minus 0.4em\relax IEEE, 2019, pp.
  370--378.

\bibitem{shawabka2020exposing}
A.~Al-Shawabka, F.~Restuccia, S.~D'Oro, T.~Jian, B.~C. Rendon, N.~Soltani,
  J.~Dy, K.~Chowdhury, S.~Ioannidis, and T.~Melodia, ``{Exposing the
  Fingerprint: Dissecting the Impact of the Wireless Channel on Radio
  Fingerprinting},'' \emph{Proc. of IEEE Conference on Computer Communications
  (INFOCOM)}, 2020.

\bibitem{Ajagannath6G2020}
A.~Jagannath, J.~Jagannath, and T.~Melodia, ``{Redefining Wireless
  Communication for 6G: Signal Processing Meets Deep Learning with Deep
  Unfolding},'' \emph{IEEE Transactions on Artificial Intelligence}, vol.~2,
  no.~6, pp. 528--536, 2021.

\bibitem{alexnet}
A.~Krizhevsky, I.~Sutskever, and G.~E. Hinton, ``Imagenet classification with
  deep convolutional neural networks,'' in \emph{Proc. of the 25th
  International Conference on Neural Information Processing Systems - Volume
  1}, ser. NIPS 12, Red Hook, NY, USA, 2012, pp. 1097--1105.

\bibitem{grc}
E.~Blossom, ``Gnu radio: Tools for exploring the radio frequency spectrum,''
  \emph{Linux J.}, vol. 2004, no. 122, p.~4, Jun. 2004.

\bibitem{lstm_watermark}
A.~Ferdowsi and W.~Saad, ``Deep learning for signal authentication and security
  in massive internet-of-things systems,'' \emph{IEEE Transactions on
  Communications}, vol.~67, no.~2, pp. 1371--1387, 2019.

\bibitem{wifi_rff}
T.~D. Vo-Huu, T.~D. Vo-Huu, and G.~Noubir, ``Fingerprinting wi-fi devices using
  software defined radios,'' in \emph{Proc. of the 9th ACM Conf. on Security
  \&; Privacy in Wireless and Mobile Networks}, 2016.

\bibitem{uav_wifi}
I.~Bisio, C.~Garibotto, F.~Lavagetto, A.~Sciarrone, and S.~Zappatore,
  ``Unauthorized amateur uav detection based on wifi statistical fingerprint
  analysis,'' \emph{IEEE Communications Magazine}, vol.~56, no.~4, 2018.

\bibitem{merchant}
K.~Merchant, S.~Revay, G.~Stantchev, and B.~Nousain, ``Deep learning for rf
  device fingerprinting in cognitive communication networks,'' \emph{IEEE
  Journal of Selected Topics in Signal Processing}, vol.~12, no.~1, 2018.

\bibitem{gtid}
S.~V. Radhakrishnan, A.~S. Uluagac, and R.~Beyah, ``Gtid: A technique for
  physical device and device type fingerprinting,'' \emph{IEEE Trans. on
  Dependable and Secure Computing}, vol.~12, no.~5, pp. 519--532, 2015.

\bibitem{AJagannath21ICC}
A.~Jagannath and J.~Jagannath, ``{Multi-task Learning Approach for Automatic
  Modulation and Wireless Signal Classification},'' in \emph{Proc. of IEEE
  Intl. Conf. on Communications (ICC)}, Canada, June 2021.

\bibitem{nmt}
D.~Bahdanau, K.~Cho, and Y.~Bengio, ``Neural machine translation by jointly
  learning to align and translate,'' \emph{ArXiv}, vol. 1409, 09 2014.

\bibitem{cv_attn1}
H.~Xu and K.~Saenko, ``Ask, attend and answer: Exploring question-guided
  spatial attention for visual question answering,'' in \emph{Proc. of Computer
  Vision -- European Conference on Computer Vision}, B.~Leibe, J.~Matas,
  N.~Sebe, and M.~Welling, Eds.\hskip 1em plus 0.5em minus 0.4em\relax Cham:
  Springer International Publishing, 2016, pp. 451--466.

\bibitem{cv_attn2}
K.~Xu, J.~L. Ba, R.~Kiros, K.~Cho, A.~Courville, R.~Salakhutdinov, R.~S. Zemel,
  and Y.~Bengio, ``Show, attend and tell: Neural image caption generation with
  visual attention,'' in \emph{Proc. of the 32nd International Conference on
  Machine Learning (ICML) - Volume 37}, ser. ICML'15.\hskip 1em plus 0.5em
  minus 0.4em\relax JMLR.org, 2015, p. 2048–2057.

\bibitem{tf_attn}
S.~Lin, Y.~Zeng, and Y.~Gong, ``Learning of time-frequency attention mechanism
  for automatic modulation recognition,'' \emph{IEEE Wireless Communications
  Letters}, vol.~11, no.~4, pp. 707--711, 2022.

\bibitem{luong-etal}
\BIBentryALTinterwordspacing
T.~Luong, H.~Pham, and C.~D. Manning, ``Effective approaches to attention-based
  neural machine translation,'' in \emph{Proc. of the Conference on Empirical
  Methods in Natural Language Processing}.\hskip 1em plus 0.5em minus
  0.4em\relax Lisbon, Portugal: Association for Computational Linguistics, Sep.
  2015, pp. 1412--1421. [Online]. Available:
  \url{https://aclanthology.org/D15-1166}
\BIBentrySTDinterwordspacing

\bibitem{vaswani}
A.~Vaswani, N.~Shazeer, N.~Parmar, J.~Uszkoreit, L.~Jones, A.~N. Gomez,
  {\L}.~Kaiser, and I.~Polosukhin, ``Attention is all you need,''
  \emph{Advances in neural information processing systems}, vol.~30, 2017.

\end{thebibliography}
\bibliographystyle{IEEEtran}


\end{document}